\pdfoutput=1

\documentclass[11pt]{article}

\usepackage[preprint]{acl}

\usepackage{times}
\usepackage{latexsym}

\usepackage[T1]{fontenc}

\usepackage[utf8]{inputenc}

\usepackage{microtype}

\usepackage{inconsolata}

\usepackage{graphicx} %
\usepackage{natbib}  %
\usepackage{caption} %
\usepackage{algorithm}
\usepackage{listings}
\usepackage{algorithmic}
\usepackage{amsmath}
\usepackage{booktabs}
\usepackage{multirow}
\usepackage[outdir=./]{epstopdf}
\usepackage{enumitem}
\usepackage{caption}
\usepackage{subcaption}
\usepackage{subfloat}
\usepackage{newfloat}
\usepackage{graphicx}
\usepackage{svg} 
\usepackage[normalem]{ulem}
\usepackage{framed}
\usepackage{mdframed}
\usepackage{xcolor}
\usepackage{lipsum}
\usepackage{float}
\usepackage{hyperref}
\usepackage{amssymb}  
\usepackage{geometry} 
\definecolor{shadecolor}{gray}{0.9}

\newlist{todolist}{itemize}{2}
\setlist[todolist]{label=$\square$}
\usepackage{pifont}
\newcommand{\xmark}{\ding{55}}%

\usepackage{array}
\newcolumntype{L}[1]{>{\raggedright\let\newline\\\arraybackslash\hspace{0pt}}m{#1}}
\newcolumntype{C}[1]{>{\centering\let\newline  \\\arraybackslash\hspace{0pt}}m{#1}}
\newcolumntype{R}[1]{>{\raggedleft\let\newline \\\arraybackslash\hspace{0pt}}m{#1}}

\AtBeginDocument{%
  \providecommand\BibTeX{{%
    \normalfont B\kern-0.5em{\scshape i\kern-0.25em b}\kern-0.8em\TeX}}
}

\title{Understanding and Evaluating Hallucinations in 3D Visual Language Models}
\author{
    Ruiying Peng\thanks{Co-first author.}, 
    Kaiyuan Li\footnotemark[1],  
    Weichen Zhang, 
    Chen Gao, 
    Xinlei Chen, 
    Yong Li \\ 
    Tsinghua University
}
\begin{document}
\maketitle
\begin{abstract}
Recently, 3D-LLMs, which combine point-cloud encoders with large models, have been proposed to tackle complex tasks in embodied intelligence and scene understanding. In addition to showing promising results on 3D tasks, we found that they are significantly affected by hallucinations. For instance, they may generate objects that do not exist in the scene or produce incorrect relationships between objects. To investigate this issue, this work presents the first systematic study of hallucinations in 3D-LLMs. We begin by quickly evaluating hallucinations in several representative 3D-LLMs and reveal that they are all significantly affected by hallucinations. We then define hallucinations in 3D scenes and, through a detailed analysis of datasets, uncover the underlying causes of these hallucinations. We find three main causes: (1) Uneven frequency distribution of objects in the dataset. (2) Strong correlations between objects. (3) Limited diversity in object attributes. Additionally, we propose new evaluation metrics for hallucinations, including Random Point Cloud Pair and Opposite Question Evaluations, to assess whether the model generates responses based on visual information and aligns it with the text's meaning.

\end{abstract}

\section{Introduction}

Large Language Models (LLMs) have demonstrated remarkable capabilities in solving complex tasks such as code completion~\cite{kanade2020learning,wang2021codet5}, mathematical reasoning~\cite{jiang2024forward,guo2024exploring}, and dialogue generation~\cite{li2024dialogue, le2020uniconv}. The success of LLMs has inspired researchers to extend this approach to more modalities, aiming to enhance models’ understanding of the real world. Large Vision-language Models, which process both images and text ~\cite{wang2024qwen2, deitke2024molmo} enabling models to interpret visual content. However, 2D information typically provides single-perspective view, which limits models' ability to accurately understand spatial relationships within a scene. Recently, researchers have explored the integration of 3D point cloud data as an input modality to improve large models’ spatial understanding. Previous works such as 3D-LLM~\cite{hong20233d-3dllm}, PointLLM~\cite{xu2024pointllm}, 3D-VLA~\cite{zhen2024-3d-vla} employed an encoder to extract features from point clouds, followed by a projector that maps these features into the tokenized space of the LLM. After fine-tuning on various 3D scene, 3D LLMs demonstrate promising performance on spatial reasoning tasks.  



Despite the success achieved by multi-modal LLMs, \textbf{hallucination}~\cite{rohrbach2018object, li2023evaluating, hu2023ciem, guan2024hallusionbench} has been identified in both Large Language Models and Large Vision-Language models. Hallucination refers to the generation of content by large models that appears correct in form but contains errors or misleading information. This poses significant challenges to the application of large models in high-risk domains such as Healthcare, Law, or Finance. Numerous methods for evaluating and detecting hallucinations have been proposed to enhance the faithfulness of large models. TruthfulQA and HalluQA have been introduced to identify hallucinations in Large Language Models (LLMs), while CHAIR and POPE have been developed to detect object-centric hallucinations in Large Vision-Language Models (LVLMs).
However, there is a lack of systematic studies on the existence of hallucinations in 3D-LLMs. In this work, we conduct a comprehensive investigation into the hallucination issue in 3D-LLMs and analyze the underlying causes.

With the introduction of the point cloud modality and depth information, the challenges faced by 3D-LLMs have shifted from image description to the accurate understanding of spatial relationships. Defining hallucinations in 3D scenes precisely has become a significant challenge. Therefore, we first define 3D hallucinations from a broad to a specific perspective, comparing them with text and visual hallucinations. Then, we use traditional evaluation method to evaluate 3D hallucinations in different representative 3D-LLMs and find they all suffer from severe hallucinations. To find the causes of severe hallucinations, we approach this analysis from the perspectives of modality co-occurrence in the training datasets and data distribution. We emphasize excessively high co-occurrence frequencies between objects is the main reason. Finally, previous 2D hallucination detection methods have predominantly focused on object-centric hallucinations, as the relationships between objects are relatively straightforward. However, in 3D scenes, there exists a multitude of complex spatial relationships. To address this, we propose a novel method and dataset capable of efficiently and accurately detecting spatial relationship hallucinations in models.

The contributions of our paper can be summarized as follows: (1) To the best of our knowledge, we are the first to investigate and define 3D hallucinations. 
(2) We evaluate several state-of-the-art 3D-LLMs and provide a detailed analysis of the causes of hallucinations.
(3) We construct a new dataset and established a detection benchmark to efficiently and accurately identify 3D hallucinations.

\begin{figure*}[t] 
\centering
\includegraphics[width=\textwidth]{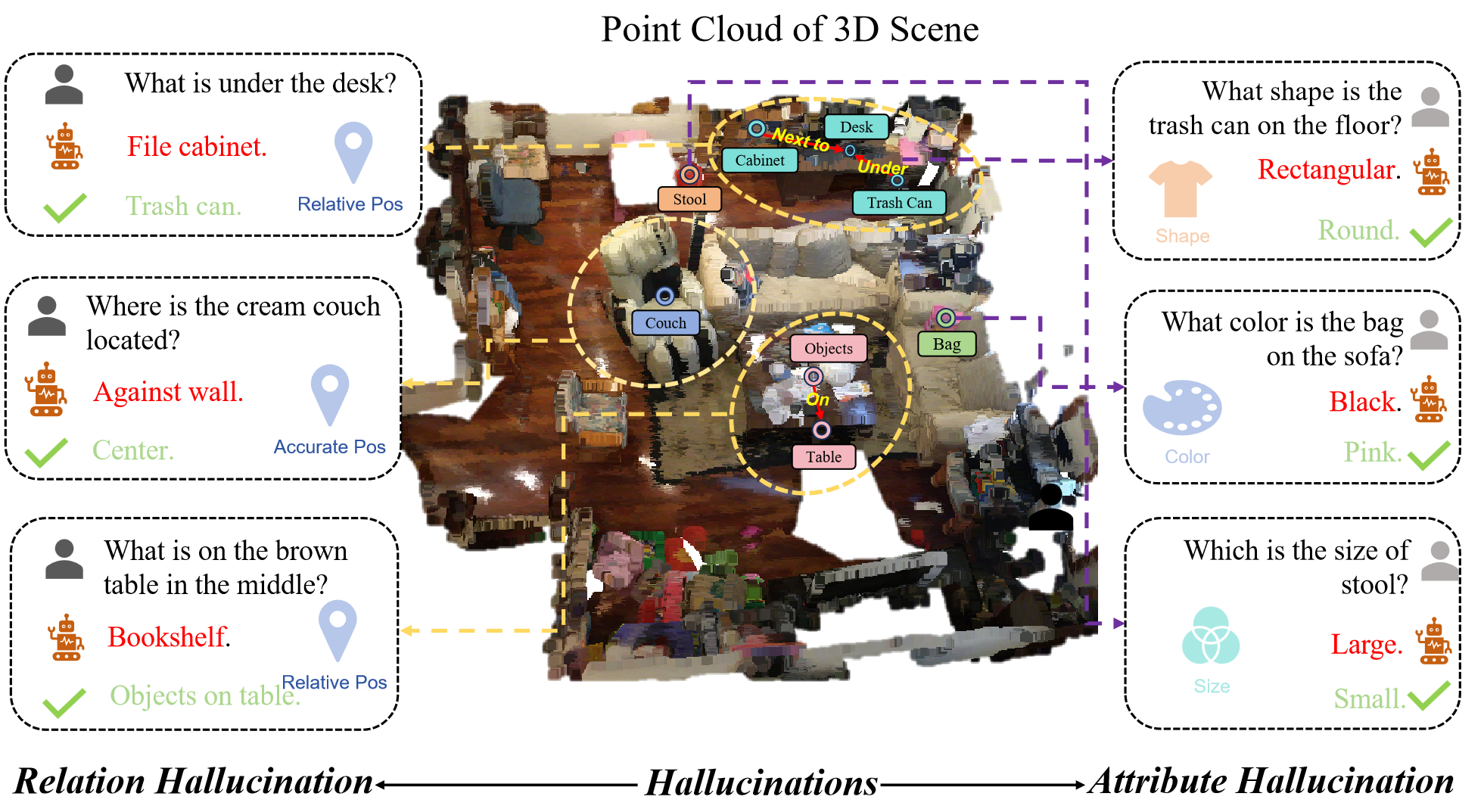} 
\caption{In 3D scenes, the relationships between objects are significantly more complex than those in text or images. The left side of the figure illustrates hallucinations related to relative positional relationships and absolute positional relationships, while the right side demonstrates attribute hallucinations such as color, size, and shape.}
\label{fig:example}
\end{figure*}
\section{Related Work}
\label{sec::related}
\subsection{3D LLM}

As Large Vision Models (LVMs)~\cite{shen2024aligning, zhang2022dino, kirillov2023segment, oquab2023dinov2} have demonstrated impressive capabilities in many tasks, extending these capabilities to other modalities has been attractive. 3D tasks, such as semantic navigation~\cite{zheng2024towards, huang2023embodied} and embodied intelligence~\cite{jatavallabhula2023conceptfusion, hong2024multiply}, have garnered widespread attention due to their relevance to real-world tasks. Many studies leverage the reasoning abilities of large language models to tackle these tasks.

In 3D-LLMs~\cite{hong20233d-3dllm}, there is typically a 3D encoder that maps the point cloud to the LLM's text language space, along with a pre-trained LLM backbone. Different models employ various methods to extract features from the point cloud.
The 3D-LLM~\cite{hong20233d-3dllm} first collects multi-view images from the scene to extract dense 2D features, and then constructs 3D features using three traditional methods; LL3DA~\cite{chen2024ll3da} uses a scene encoder pre-trained on the ScanNet~\cite{dai2017scannet} detection task as the input encoder for the point cloud modality; Leo~\cite{huang2023embodied} employs an object-centric 3D token embedding method, where each object is first encoded using a pretrained point cloud encoder, and then further processed with a spatial transformer. After fine-tuning on the 3D downstream tasks, these models have shown remarkable spatial abilities.

After encoding the point cloud and fine-tuning on 3D tasks, 3D-LLMs have achieved promising results on tasks such as 3D Dense Captioning, 3D Question Answering, and Scene Description.

\subsection{Hallucination in MLLM}

In large language models, hallucinations are defined as instances where the model generates outputs that appear reasonable but are not faithful to the facts or the provided context~\cite{filippova-2020-controlled}. Hallucinations in large models have a significant impact on their practical deployment and user experience. Existing work~\cite{leng2024mitigating, liu2023mitigating, yu2024hallucidoctor, zhai2023halle} often mitigates hallucinations through methods such as model editing, post-training, and contrastive decoding. As large language models are increasingly used as backbones in multimodal fields, the hallucinations in LVLMs has garnered widespread attention from researchers. In LVLMs, hallucinations are defined as instances where the generated text response does not align with the corresponding visual content. Current research ~\cite{rohrbach2018object, li2023evaluating, hu2023ciem} mainly focuses on object-related hallucinations, which can be categorized into hallucinations concerning object types, object attributes, and object relationships. Similar to methods in LLMs, existing works often mitigate hallucinations in LVLMs through data-related, training-related, and contrastive decoding-related approaches.

Hallucinations pose serious security concerns for the application of large models in real-world scenarios, particularly for embodied intelligence and spatial navigation tasks that take 3D scenes as input. However, to the best of our knowledge, no work has yet discussed hallucinations in 3D scenes. Therefore, in this work, we address the hallucinations presented in 3D-LLMs from the perspective of detection and analysis.

\section{3D Hallucination}
\label{sec::system}
In this section, we first validate the existence of significant hallucination issues in the current popular 3D-LLMs on the 3D captioning task using traditional object-centric method which is used in image hallucination evaluate. We then define 3D hallucinations and compare them with the multimodal hallucinations defined in previous works.

\subsection{Simple Evaluation Based on Traditional Detection Methods}
\begin{table}[H]
    \centering
   \resizebox{1.0\linewidth}{!}{ \begin{tabular}{cccccc}
    \toprule
     & Precision & Recall & F1Score & Rouge & Meteor \\
    \midrule
    ll3da & 36.36 & 16.67 & 22.86 & 25.87 & 14.98 \\
    3D-LLM & 22.97 & 8.20  & 10.92 & 9.94  & 4.37  \\
    \bottomrule
    \end{tabular}}
    \caption{Evaluate Result of Sota 3D-LLM}
    \label{table:hallucinatiovalid}
\end{table}
First, we evaluate whether existing 3D-LLMs are affected by object hallucinations in tasks where it is relatively easier compare with relation hallucinations. Here, we use traditional object hallucination definitaion in image-text area which defines object hallucinations as situations where the items described in the model's output do not exist in the real scene. If the object described by 3D-LLM mismatch with the ground truth, we consider that a hallucination has occurred. Formally, we define \( A \) as the set of items output by the model, and \( B \) as the set of items present in the real scene. The evaluation metric can be defined as:
 \begin{align}
      Precision= \frac{|A \cap B|}{|A|} 
\end{align}
 \begin{equation}
      Recall= \frac{|A \cap B|}{|B|} 
 \end{equation}
 
\begin{figure}[H]
\begin{center}
\begin{minipage}{0.9\linewidth}
    \includegraphics[width=1.0\linewidth]{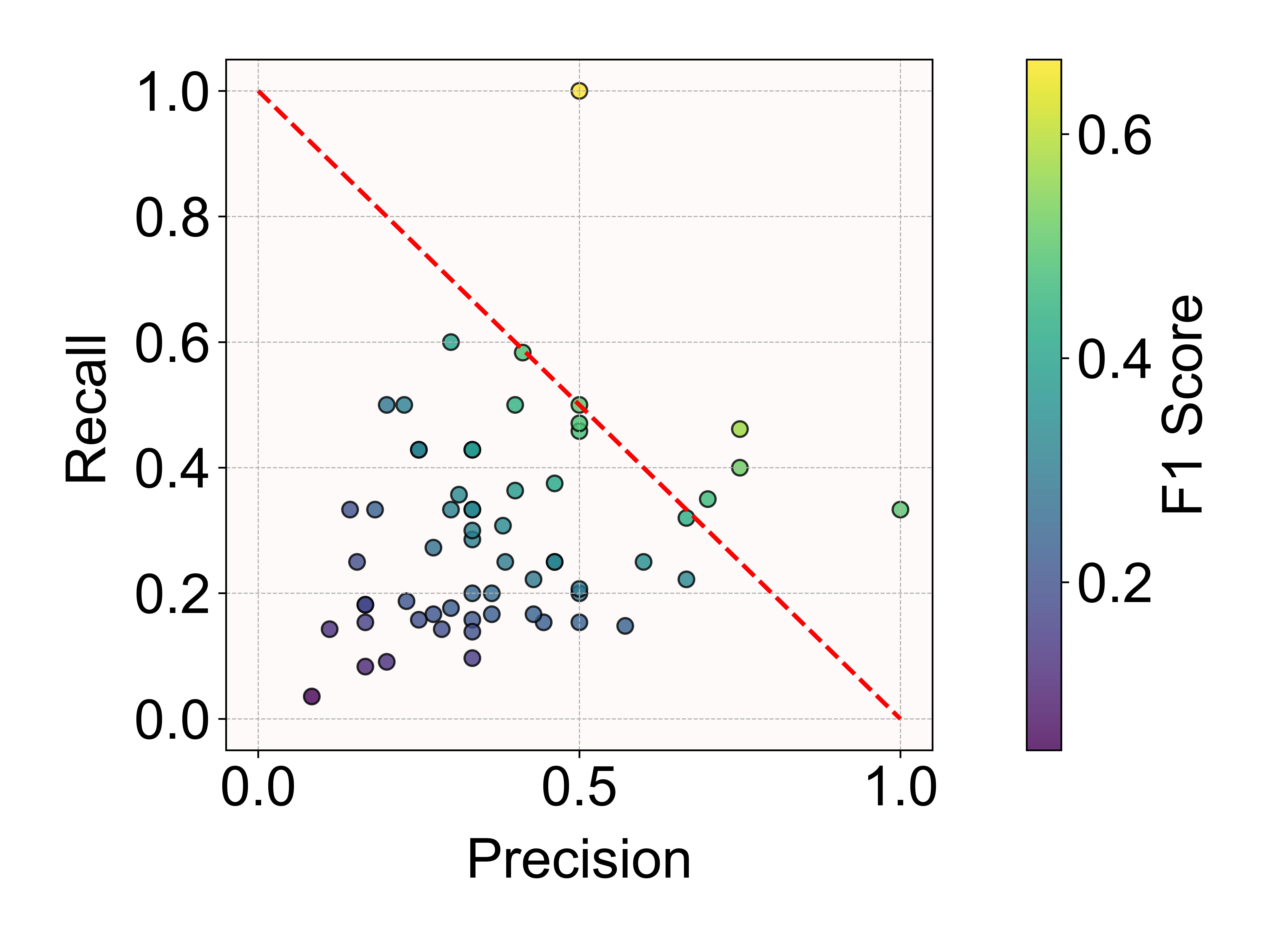}
     \caption{Object hallucination evaluation for 3D LLMs. Precision measures the proportion of described objects that exist in the scene, while recall represents the proportion of scene objects that are described.}
     \label{fig:hallucination_eva}
\end{minipage}
\end{center}
\end{figure}

To validate that existing 3D models suffer from significant object hallucinations, we select two representative 3D models : ll3da and 3D-LLM for evaluation. We use the metric defined above. The results are presented in Table \ref{table:hallucinatiovalid}. As we can see, both models perform badly and exhibit significant hallucination issues in the object description task. To better illustrate the evaluation of hallucinations, we present our evaluation of LL3DA on the description task as a Recall-Precision plot, as shown in Fig. \ref{fig:hallucination_eva}. The image is divided into the bottom-left corner and the top-right corner. The bottom-left corner indicates that the model struggles with hallucinations in the object description task, while the top-right corner demonstrates that the model performs well.  It can be observed that most of the samples are concentrated in the lower-left corner of the plot, which reflects the presence of severe hallucinations in the majority of examples produced by the current state-of-the-art models.

\subsection{3D Hallucination Definition}

\subsubsection{Modality Difference}
Previous work on hallucinations has primarily focused on text and image modalities, as well as their interactions. Since the main difference between 3D-LLMs and earlier vision-based large models lies in the input modalities, we approach the analysis and comparison of the three types of hallucinations from the perspective of input modalities.
As shown in Table \ref{table:Different_modality}, unlike text-based LLMs and text-image-based LVLMs, 3D-LLMs primarily use text and point cloud modalities as inputs, which brings extra depth information.

\begin{table}[h]
\centering
\Large
\resizebox{1.0\linewidth}{!}{
\begin{tabular}{ccccccc}
\toprule
\multirow{3}{*}{Model Type} & 
\multicolumn{3}{c}{Input Modality} &
\multicolumn{3}{c}{Modality Conflict} \\ \cmidrule(lr){2-7}
& Text    & Vision   & Depth   & Knowledge Conflict & Text-Image Conflict & Scene Conflict \\ \midrule
\multirow{1}{*}{LLM} & \checkmark & \xmark  & \xmark  & \checkmark & \xmark & \xmark \\
\multirow{1}{*}{LVLM} & \checkmark & \checkmark & \xmark & \checkmark & \checkmark & \xmark  \\
\multirow{1}{*}{3D-LLM} & \checkmark & \checkmark & \checkmark & \checkmark & \checkmark & \checkmark  \\
\bottomrule
\end{tabular}}
\caption{Modality Difference}
\label{table:Different_modality}
\end{table}

\begin{table}[h]
\centering
\Large
\resizebox{1.0\linewidth}{!}{
\begin{tabular}{ccccccc}
\toprule
\multirow{3}{*}{Model Type} & 
\multicolumn{3}{c}{Object Hallucination} &
\multicolumn{3}{c}{Relation Hallucination} \\ \cmidrule(lr){2-7}
&Color & Shape & Size & Abstract& Relative & Accurate \\ \midrule
\multirow{1}{*}{Text Hallucination} & \checkmark & \xmark  & \xmark  & \checkmark & \xmark & \xmark \\
\multirow{1}{*}{Image Hallucination} & \checkmark & \checkmark & \xmark & \checkmark & \checkmark & \xmark  \\
\multirow{1}{*}{3D Hallucination} & \checkmark & \checkmark & \checkmark & \checkmark & \checkmark & \checkmark  \\
\bottomrule
\end{tabular}}
\caption{Classification of Hallucinations}
\label{table:classification}
\end{table}

The uniqueness of the input modalities leads to differences in the interactions between modalities. In text hallucinations, conflicts only arise between different textual knowledge, i.e., knowledge conflicts, which are also presented in LVLMs and 3D-LLMs, as both are built on LLMs. In image hallucinations, conflicts occur between textual and visual information. However, in 3D hallucinations, the depth information leads to conflicts where 3D-LLMs generates fictitious spatial relationships within the scene. We refer to this phenomenon as \textbf{scene conflict}.

\subsubsection{Hallucination Definition}

To define hallucination types appeared in scene conflict more concretely and accurately, we abstract the 3D scene into objects and relationships, thus defining two types of hallucinations: \textbf{Object hallucinations} and \textbf{Relation hallucinations}. We present the classification in Table \ref{table:classification}.

Object hallucinations are primarily related to the attributes of objects, such as color, shape, and size. Among these attributes, \textbf{size attribute} requires accurate depth information for proper evaluation, making this a hallucination type unique to 3D scenes. Formally, we use \(H_{obj}\) to represent object hallucination, \(S\) to represent the attributes set. \(Attr^{i}_{true} \in S\) represents the real object's attribute. \(Attr^{i}_{pred}\) represents the attributes in the prediction of 3D-LLM. 

\begin{equation}
    H_{obj} = S[Attr^{i}_{true} \neq Attr^{i}_{pred}]
\end{equation}

Relation hallucinations, on the other hand, are primarily concerned with the relationships between objects.
Among these relations, \textbf{Abstract relationship hallucinations} refer to the functional relationships between objects. \textbf{Relative positional relationships} refer to broader postion relationships, such as left-right orientation, which can usually be inferred from a given view. However, because a single view lacks depth information, precise positional relationships, such as "hanging" or "standing on," cannot be determined. In 3D scene, we can deduce \textbf{accurate spatial relations} among objects. Formally, we use \(O_i\) and \(O_j\) to represent two objects, use \(\stackrel{rel}{\longrightarrow}\) to represent relationship between two objects, use \(\stackrel{pred}{\longrightarrow}\) to represent predicted relationship. The we can define relation hallucination as:

\begin{equation}
    O_i \stackrel{rel}{\longrightarrow} O_j \neq O_i \stackrel{pred}{\longrightarrow} O_j
\end{equation}

\section{Data Bias Intensifies 3D Large Model Hallucinations}\label{sec::system}
\begin{figure*}[htb] 
\centering
\includegraphics[width=\textwidth]{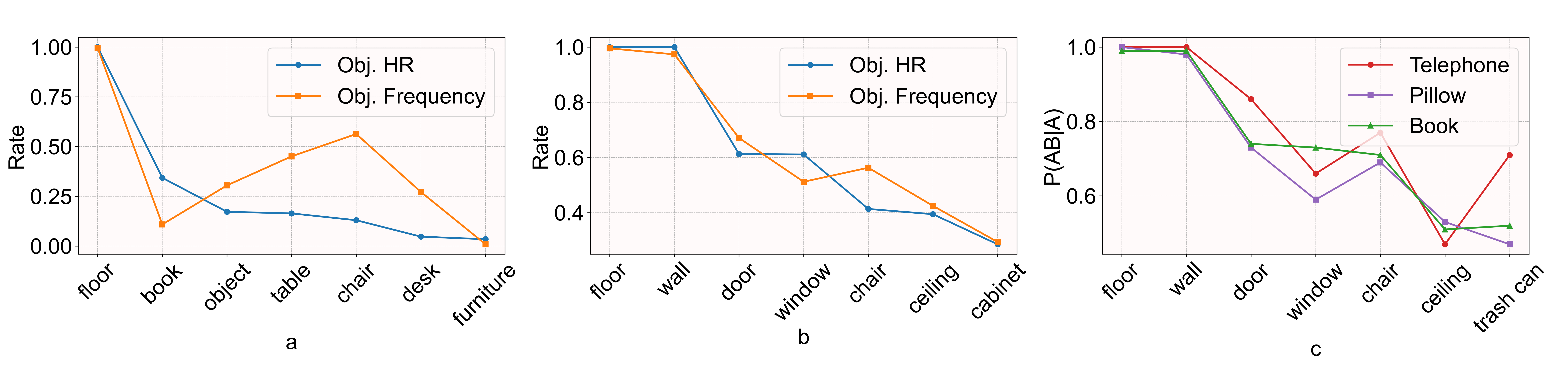} 
     \caption{(1): The relationship between object hallucination rates in 3DLLM and LL3DA and object occurrence frequencies in the dataset is shown in figures a and b.(2): The relationship between strong object correlations and object hallucination rates are shown in figure c.}
     \label{fig:influenceOfDataset}
\end{figure*}
In the previous section, we briefly examined the significant hallucinations present in existing 3D large models and provided an analysis and definition of hallucinations in 3D scenes. In this section, we will delve into the underlying causes of this phenomenon.
In Section 3 of our study, we evaluated the occurrence of object hallucination in  large 3D point cloud models. We found that  model often describes objects that do not exist in the actual scene.We hypothesize that imbalanced object frequencies and object corelation in the dataset contribute heavily to hallucination. 
\subsection{Imbalanced Frequency Distribution of Objects}
We performed statistical analysis on the hallucination rate and occurrence frequency of objects. The hallucination rate$(HR)$ of an object is defined as the ratio of scenes in which the object is incorrectly identified as present, even though it does not actually exist, to the total number of scenes where the object is absent in the test set. The occurrence frequency of an object is defined as the ratio of scenes where the object is present to the total number of scenes. As shown in Figure \ref{fig:influenceOfDataset}, $a$ represents the object hallucination rate results for 3DLLM, and $b$ represents the object hallucination rate results for LL3DA. From the figures, it can be observed that the curve representing the hallucination rate closely follows the curve representing the occurrence frequency. This suggests that objects with a high frequency of occurrence are more likely to be accurately described by the model, as it tends to repeat the most common elements. In other words, \textbf{objects with higher occurrence frequencies are more prone to hallucination}, being more likely to be incorrectly identified as present when they are actually absent.

However,in the Scannet dataset, certain objects such as the floor, wall, and door appear very frequently across many scenes.\textbf{Floor} appeared in \textbf{1506 out of 1513} scenes.\textbf{Wall} appeared in \textbf{1473} scenes.\textbf{Door} appeared in \textbf{1015} scenes.These data demonstrate that scene similarity in ScanNet is high, with the same object appearing repeatedly across multiple scenes.Based on the conclusion that excessively high occurrence frequencies can exacerbate hallucinations, we can infer that \textbf{the high overlap of objects across different scenes in the dataset} is one of the key factors \textbf{contributing to the strong hallucinations} observed in 3D large language models.
\subsection{Potential Influence of Object Correlation}
 In Figure \ref{fig:influenceOfDataset}, the y-axis represents the conditional probability \( P(AB|A) \), where A denotes the presence of object a in the scene and B denotes the presence of object b. A higher value of \( P(AB|A) \) indicates a higher likelihood that if object a is present, object b is also likely to be present. The objects b labeled on the x-axis, such as floor, wall, and door, are arranged in descending order of their hallucination rates, and the conditional probabilities also exhibit a downward trend. This suggests that \textbf{objects frequently co-occurring with others are more likely to be incorrectly identified as present}, thereby inducing hallucinations. For example, if \textbf{chairs} and \textbf{tables} often appear together in the same scene, the model might learn an implicit dependency between them. When the chair is present, the model may "hallucinate" the table, even if it isn't present in the actual scene.\\
 ScanNet is an indoor scene dataset containing environments such as bedrooms, bathrooms, and offices. Due to the specific nature of these scenes, they consistently include certain objects—such as toilets, sinks, and toilet paper—always appearing together in bathrooms. This strong correlation between objects in the dataset means that during training, the model may receive rewards for providing answers based on these associations rather than point clouds. As a result, the model may incorrectly associate these objects with one another, leading to hallucinations when detecting one object.

\begin{figure*}[t] 
\centering
\includegraphics[width=\textwidth]{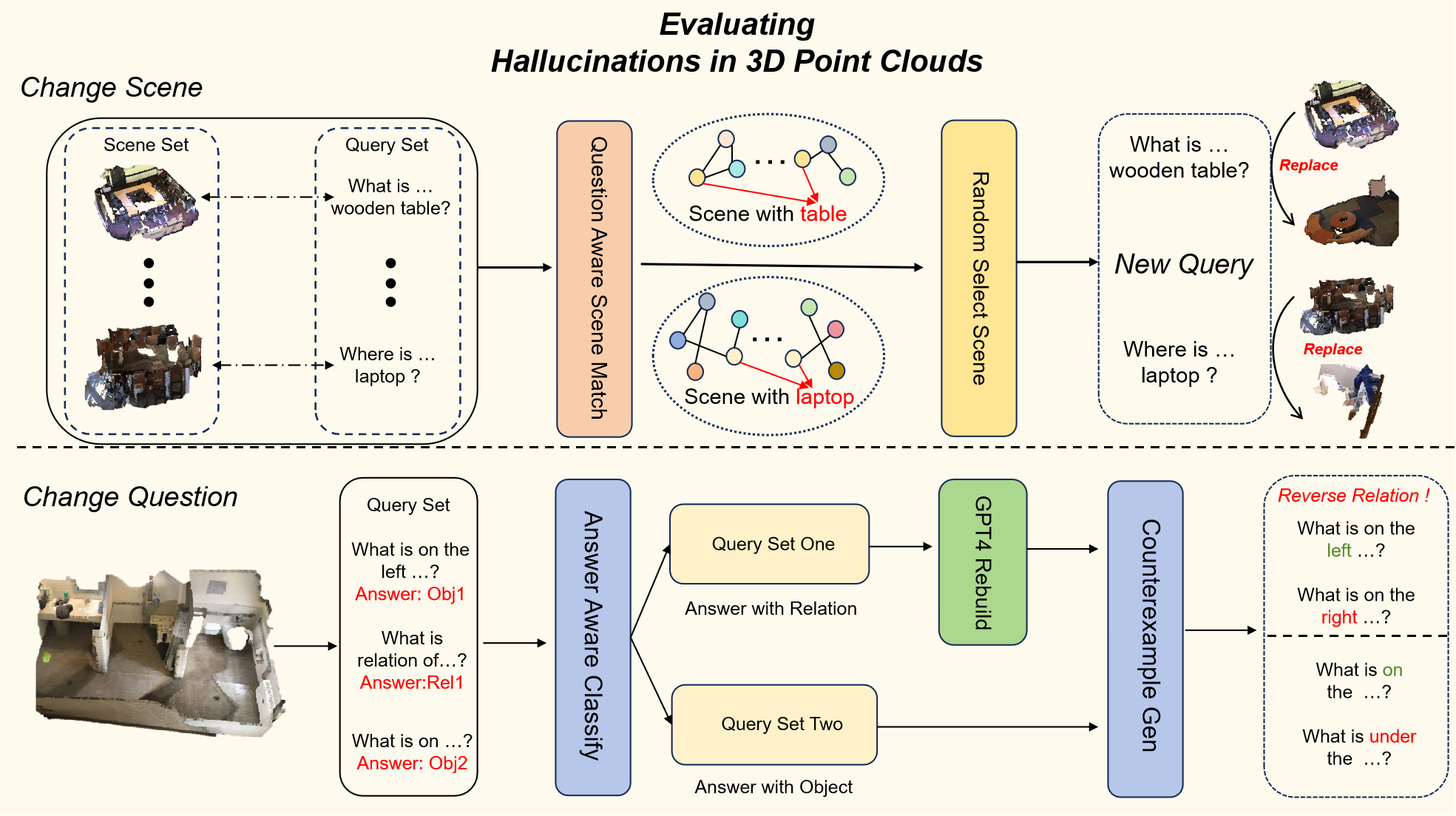} 
     \caption{In the evaluation process, we generate new QA pairs by changing the scene while keeping the questions fixed: different scenes are randomly selected to form new QA pairs. Additionally, we modify the questions while keeping the scene fixed: spatial relationship-related questions are selected, and all QA pairs are transformed such that the object A is the focus. Then, the spatial relationship in the questions is inverted, generating new QA pairs.}
     \label{fig:evaluationProcess}
\end{figure*}

\makeatletter
\renewcommand{\subsubsection}[1]{%
  \par\noindent\textbf{#1}\par 
}
\makeatother

\section{Evaluation and Detection}\label{sec::exp}


\subsection{ Inadequacy of Existing Evaluation Frameworks}
Existing evaluation frameworks for 2D multimodal models, such as POPE~\cite{li2023evaluating}, are insufficient for addressing the challenges in 3D point cloud large language models (LLMs). Since the POPE view uses yes/no questions to evaluate model object hallucinations, which cannot accurately assess the model's understanding of spatial relationships or visual details such as attributes.In Section 3, we assess hallucinations in 3D point cloud models by evaluating object hallucination in description tasks. However, this method has two main limitations: 1) It only detects hallucinations in description tasks, as not all responses involve objects. 2) It doesn't analyze other types of hallucinations, such as attribute or relational errors.

 Therefore, we aim to propose a more stable, fair, and flexible evaluation framework for evaluating hallucinations in 3D point clouds.
\begin{table*}[h]
\centering
\small
\begin{tabular}{llccc|ccc}
\hline
\multicolumn{2}{c} {\multirow{3}{*}{Type}} & \multicolumn{3}{c|}{ll3da} & \multicolumn{3}{c}{3dllm} \\ \cline{3-8}
 
     &   & \multicolumn{2}{c}{Accuracy} & \multirow{2}{*}{\textbf{$HR_{random}$\%}}  & \multicolumn{2}{c}{Accuracy} & \multirow{2}{*}{\textbf{$HR_{random}$\%}}  \\
     
     &   & Rouge-L & Meteor&  & Rouge-L & Meteor&  \\ \hline

\multirow{4}{*}{Relation} & Direction  & 30.62  & 19.53 &\textbf{33.21} & 30.32  & 19.77 &\textbf{30.43}\\
                              & Containment  & 43.28  & 35.27&\textbf{36.89}  & 42.51  & 31.98&\textbf{43.69} \\ 
                              & Contact & 35.08  & 23.55 &\textbf{34.72}  & 35.58  & 24.2 &\textbf{36.79}\\ 
                              & Distance    & 32.02  & 22.71 &\textbf{31.49} & 32.5  & 21.36&\textbf{28.94} \\ \hline
                             
\multirow{3}{*}{Property}  & Color                    & 47.38  & 41.9 &\textbf{62.69} & 51.72   & 47.38&\textbf{61.77} \\ 
                                      & Shape                    & 42.74  & 31.9  &\textbf{49.48}& 44.56  & 32.94 &\textbf{46.39}\\ 
                                      & Size                     & 43.74  & 39.01 &\textbf{74.29} & 47.48  & 37.57&\textbf{51.43} \\ \hline
                                     
\multicolumn{2}{c}{Comparison}                 & 24.75  & 17.65 &\textbf{63.16} & 29.43  & 21.82&\textbf{42.11} \\ 
\multicolumn{2}{c}{Quantity} & 50.18  & 41.84&\textbf{63.93}  & 49.85  & 42.68 &\textbf{53.88} \\ 
\multicolumn{2}{c}{Usage} & 32.22  & 21.62 &\textbf{34.78} & 30.88  & 22.03&\textbf{26.09} \\ 
\multicolumn{2}{c}{Other}  & 37.22  & 31.72&\textbf{38.89}  & 39.8  & 32.82&\textbf{25.93} \\ 
\hline
\end{tabular}
\caption{Model Performance and Hallucination Rate in Random Scenarios.Accuracy refers to the evaluation result between the model's response and the ground truth. $HR_{random}$ is the hallucination rate calculated based on random scenes as defined in Section 5.}
\label{table:randomSceneResult}
\end{table*}
\begin{figure*}[h]
\begin{center}
\centering
    \includegraphics[width=0.8\linewidth]{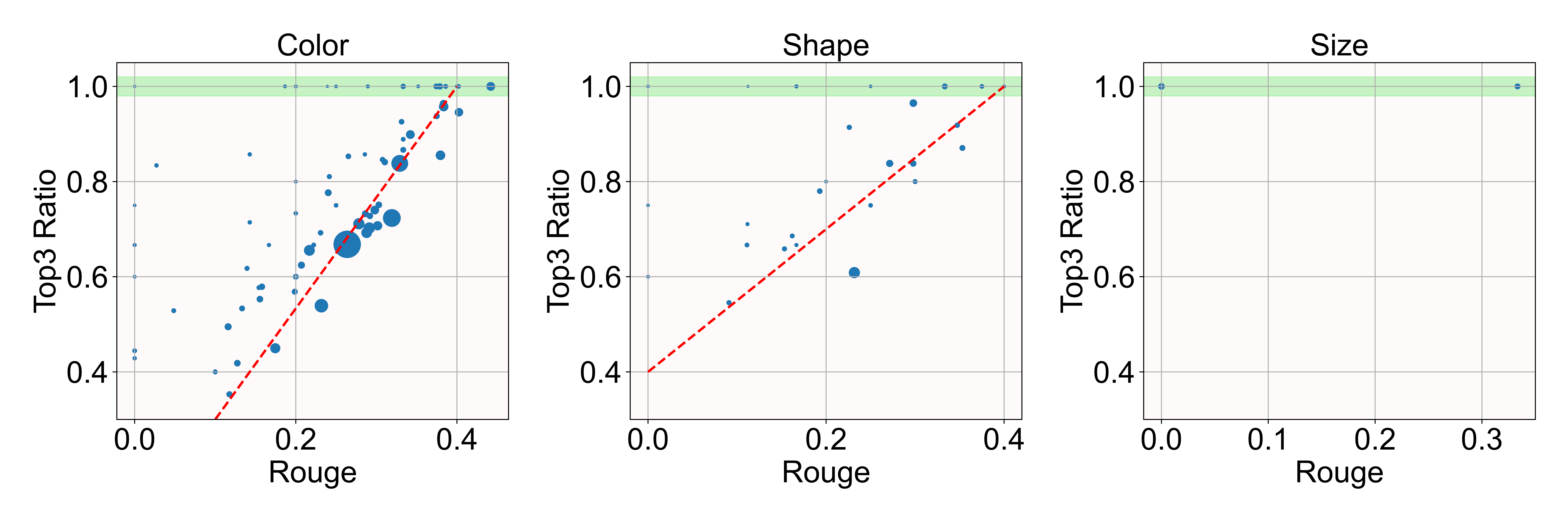}
     \caption{Impact of Attribute Simplicity on Accuracy.ROUGE represents the average quality of question-answer pairs for a specific item, while the Top 3 Ratio is the proportion of the three most common attributes of the item.}
     \label{fig:topkRatio}
\end{center}
\end{figure*}
\subsection{Proposed Evaluation Framework}
We propose two strategies for detecting hallucinations in 3D point cloud models.\\
\textbf{\emph{Random Point Cloud Pair Evaluation}} ~We select a random point cloud and ask the model the same question on both the original and new point clouds. If the answers are identical, it's considered a hallucination, indicating the model doesn't integrate visual context and just maps the question to a fixed answer.\\
\textbf{\emph{Opposite Question Evaluation}} ~For a fixed point cloud, we ask two Opposite questions (e.g., "What is on the right of the table?" and "What is on the left?"). If the model gives the same answer, it's a hallucination, suggesting the model isn't using the spatial information from the point cloud.

 By employing these two strategies, we aim to identify cases where the model fails to distinguish between spatially different scenarios or produces inconsistent responses to questions.
\subsection{ Inadequacy of Existing Evaluation Frameworks}
The entire pipeline is illustrated in Figure \ref{fig:evaluationProcess}. \\
\textbf{Data Generation}: In the \textbf{change scene} experiment, for each $(Q_i,A_i,S_i)$ pair, we randomly select a different $S_j$ from the scene set to create a new $(Q_i,A_i,\{S_i, S_j\})$ pair dataset.In the \textbf{change question} experiment, we first select questions related to spatial relationships and use GPT-4 to convert each QA pair into a dataset where the answer is an object, resulting in the \textit{scanqa-SR} dataset. For each spatial relationship question $Q_i$ in \textit{scanqa-SR}, we generate its opposite $Q_j$ to form $(\{Q_i, Q_j\},A_i,S_i)$ pairs, creating the \textit{scanqa-SR-Opposite} dataset.\\
\textbf{Experiment}: We then conduct tests using the aforementioned data on different models. 
In Experiment 1, for a given question \( q_i \), we generate two answers, \( a_{ij} \) and \( a_{ik} \), corresponding to two different scenes, \( s_{j} \) and \( s_{k} \), respectively.
We use BLEU-4~\cite{papineni2002bleu}, ROUGE~\cite{lin-2004-rouge}, and METEOR~\cite{banerjee2005meteor} metrics to measure the similarity between two answers. The hallucination rate($HR_{random}$) is calculated as follows:
\begin{equation}
HR_{\text{random}} = \frac{1}{N} \sum_{i=1}^{N} \mathbf{1}(\text{ROUGE}(a_{ij}, a_{ik}) > 0.66)
\end{equation}
In Experiment 2, for a fixed scene \( s_i \), we generate two answers, \( a_{ji} \) and \( a_{jk} \), for two semantically opposite questions, \( q_j \) and \( q_k \).The hallucination rate($HR_{opposite}$) is calculated as follows:
\begin{equation}
HR_{\text{opposite}} = \frac{1}{N} \sum_{i=1}^{N} \mathbf{1}(\text{ROUGE}(a_{ji}, a_{jk}) > 0.66)
\end{equation}

\begin{table*}[h]
\centering
\small
\begin{tabular}{lllcccc}
\hline
Dataset & Task &Model & Bleu-4  & Rouge-L & Meteor   &\textbf{$HR_{opposite}$\%}  \\ \hline
\multirow{6}{*}{scannet}&\multirow{2}{*}{scanqa}&ll3da&7.64&36.56&26.95&/\\ 
& &3dllm&0.80&37.46&28.18&/ \\  \cline{2-7}
&\multirow{2}{*}{scanqa-SR}&ll3da&0.02&13.34&9.68&/ \\ 
&&3dllm&0.0&15.55&10.28&/ \\   \cline{2-7}
&\multirow{2}{*}{scanqa-SR-Opposite}&ll3da&/&/&/&\textbf{56.27} \\ 
&&3dllm&/&/&/&\textbf{52.25} \\ \hline
\end{tabular}
\caption{Model Performance and Hallucination Rate on Semantically Opposite Questions.BLEU-4, ROUGE, and METEOR are evaluation metrics for model response quality based on ground truth, while $HR_{opposite}$ represents the hallucination rate in the opposite-question experiment.}
\label{table:oppositeQuestionResult}
\end{table*}
\section{Evaluation on 3dllm and ll3da}
\subsection{Hallucinations in Random Scene Queries}
We evaluate two models using the approach above. Table \ref{table:randomSceneResult} presents the results for random scenes. ROUGE and METEOR measure performance on ScanQA, while $HR_{random}$ is defined in Section 5.2. The table shows a positive correlation between accuracy and hallucination rate. LL3DA and 3DLLM both exhibit low accuracy and hallucination rates for spatial questions but higher rates for object attributes.\\
This suggests that the model exhibits significant hallucination issues, where it answers questions without considering the visual context, yet its responses appear 'better' or closer to the ground truth. Upon examining the training set, we find that object attributes often align with typical characteristics—for example, tables are usually black, white, or brown, and televisions are typically rectangular. This indicates that the model learns attribute associations due to the homogeneous nature of indoor scenes and the limited diversity of attributes.
\subsection{Relationship Between Attribute Uniformity and Answer Accuracy}
We plotted Figure \ref{fig:topkRatio} to illustrate the relationship between the uniformity of an object's properties and the accuracy of the answers. For instance, chair color is queried 346($N$) times, with black ($T_1$ times), brown ($T_2$ times), and gray ($T_3$ times) as the most frequent colors. To quantify attribute uniformity, we introduce the "Top-K Ratio," where the Top-3 Ratio for the chair can be calculated as: 
\begin{equation}
\text{Top-3 Ratio} = \frac{T_1 + T_2 + T_3}{N}.
\end{equation} The x-axis of the figure represents the average ROUGE score for questions related to a specific object, with higher ROUGE scores indicating that the questions regarding the object's properties are more easily answered correctly. The three plots from left to right show the relationship between the accuracy of answers and the uniformity of the object's properties, specifically color, shape, and size.
In the plots for color and shape, the distribution of points is approximately linear, confirming a strong positive correlation between the accuracy of the answers and the uniformity of the object's properties. Additionally, we observed that many points clustered near a Top-3 Ratio of 1, suggesting that the dataset contains objects with highly uniform attributes. Such objects tend to exhibit a strong correlation between the object and a specific attribute, which makes it easier for the model to hallucinate the correct attribute. 
\subsection{Hallucinations in Opposite-Question Queries}The results for testing with 
opposite questions within the same scene are presented in Table \ref{table:oppositeQuestionResult}.The ScanQA dataset includes a wide range of QA pairs involving various attributes, spatial relationships, and other data types. In contrast, ScanQA-SR focuses solely on spatial relationships and transforms all QA pairs into those where the answer is the object itself.\\
By comparing the results from these two datasets, we observe that the ROUGE scores for ScanQA-SR are significantly lower than those for ScanQA. This indicates that the model is more prone to errors when dealing with spatial relationship tasks. To investigate whether the model truly understands the meaning of spatial relationships, we created a dataset of opposite questions specifically for spatial relationships. The goal was to assess the model’s ability to handle questions about opposing spatial positions.\\
However, we found that the hallucination rate for both models exceeded 50\%. This suggests that when posed with opposite questions about the same scene, the model has a 50\% chance of giving the same answer. This further demonstrates that the model is prone to errors and hallucinations when handling spatial relationship queries. The results imply that the model may lack a proper visual-semantic understanding of spatial relationships, leading it to answer incorrectly without considering point cloud data.

\section{Conclusion}\label{sec::conclusion}
This study classifies 3D hallucinations and evaluates the severity of hallucinations in the large-scale point cloud models 3DLLM and LL3DA through description and QA tasks. By analyzing hallucination rates across datasets, we identify that high object frequency, strong correlations, and attribute singularity contribute to hallucinations. We explore whether models rely on visual information, but current tasks and metrics only measure text similarity to ground truth. To address this, we design two experiments and define hallucinations based on the results. Our findings show that models struggle to answer contextually accurate questions and align spatial relationships with visual concepts.



\section{Limitations}

In this study, we provide a detailed classification of hallucination types specifically for the QA task. Each QA pair is classified to detect corresponding hallucinations. However, for the description task and other long-text tasks, no specific approach is proposed to detect the types of hallucinations present in the generated answers. This limitation means that our evaluation only demonstrates the significant hallucination issues within 3D point cloud models, and uses different types of short QA pairs to explore the following questions: 1) Which types of questions are more likely to induce hallucinations in the model? 2) How does the dataset distribution impact the occurrence of hallucinations in the model?

Furthermore, we identify that models are particularly prone to attribute hallucinations and investigate the relationship between dataset distribution and hallucination rates. Regarding spatial relationship hallucinations, our experiments only reveal that the models lack understanding of spatial relationships, but do not explain why the models perform worse on spatial relationship-related questions compared to other question types.

Third, in our experiments designed to explore whether the models answer based on visual information or rely on textual inputs alone, the results indicate that the current dataset is overly simple and highly regular, which allows the models to disregard visual information in favor of answering based on text alone. However, we do not provide insights into why the models do not incorporate point cloud information in their responses from an architectural perspective.

Finally, we utilize GPT-4 to generate a new annotated dataset, which, compared to manual annotation, may contain some minor errors. Although we have discussed hallucination issues in 3D large language models and highlighted the problem of models not responding based on point cloud data, this should not be interpreted as a pessimistic view of the development of 3D language models. On the contrary, we aim to identify the reasons behind their suboptimal performance, such as the dataset distribution issues discussed in this paper. We hope that our work can provide new insights and ideas for further improving the performance of 3D large language models.

\bibliography{bibliography}




\end{document}